\documentclass[11pt,letterpaper]{article}
\usepackage[letterpaper]{geometry}
\usepackage{acl2012}
\usepackage{times}
\usepackage{latexsym}
\usepackage{amsmath}
\usepackage{multirow}
\usepackage{url}
\makeatletter
\newcommand{\@BIBLABEL}{\@emptybiblabel}
\newcommand{\@emptybiblabel}[1]{}
\makeatother

\usepackage{todonotes}

\newcommand{\ignore}[1]{}

\usepackage{color}
\usepackage{array}
\usepackage{amsmath,amsthm,amsfonts}
\usepackage{graphicx,caption,subcaption}
\usepackage{float}
\usepackage{algorithm}
\usepackage{algorithmic}
\usepackage{framed}
\usepackage[hidelinks]{hyperref}

\DeclareMathOperator{\Op}{\odot}

\DeclareMathOperator{\Score}{\textsc{Score}}

\DeclareMathOperator{\Coref}{\mathrm{Coref}}

\DeclareMathOperator{\Template}{\textsc{Template}}

\DeclareMathOperator{\UnitDep}{\textsc{UnitDep}}
\DeclareMathOperator{\Knowledge}{\textsc{Knowledge}}

\newcommand\blfootnote[1]{%
  \begingroup
  \renewcommand\thefootnote{}\footnote{#1}%
  \addtocounter{footnote}{-1}%
  \endgroup
}

\title{Mapping to Declarative Knowledge for Word Problem Solving}

\author{Subhro Roy$^{*}$ \\
  Massachusetts Institute of Technology \\
  {\tt subhro@csail.mit.edu} \\\And
  Dan Roth$^{*}$\\
  University of Pennsylvania \\
  {\tt danroth@seas.upenn.edu} \\}

\date{}

\begin{document}
\maketitle
\blfootnote{$^{*}$Most of the work was done when the authors were at the University of Illinois, Urbana Champaign.} 
\begin{abstract}
Math word problems form a natural abstraction to a range of
quantitative reasoning problems, such as understanding financial news,
sports results, and casualties of war. Solving such problems requires
the understanding of several mathematical concepts such as dimensional
analysis, subset relationships, etc.
In this paper, we develop declarative rules which govern the
translation of natural language description of these concepts to math
expressions. We then present a framework for incorporating such
declarative knowledge into word problem solving. Our method learns to
map arithmetic word problem text to math expressions, by learning to
select the relevant declarative knowledge for each operation of the
solution expression.
This provides a way to handle multiple concepts in the same problem
while, at the same time, support interpretability of the answer
expression.
%
%
Our method models the mapping to declarative knowledge as a
latent variable, thus removing the need for expensive annotations.
Experimental evaluation suggests that our domain knowledge based
solver outperforms all other systems, and that it generalizes better
in the realistic case where the training data it is exposed to is
biased in a different way than the test data.

\end{abstract}

\section{Introduction}
\label{sec:intro}
Many natural language understanding situations require reasoning with
respect to numbers or quantities -- understanding financial news,
sports results, or the number of casualties in a bombing. Math word
problems form a natural abstraction to a lot of these quantitative
reasoning problems. Consequently, there has been a growing interest in
developing automated methods to solve math word
problems \cite{KZBA14,HHEK14,RoyRo15}.

\begin{figure}[H]
  \centering
  \small       
  \begin{tabular}{|p{7.5cm}|} \hline
    \textbf{Arithmetic Word Problem} \\\hline

    Mrs. Hilt baked pies last weekend for a holiday dinner. She baked
    16 pecan pies and 14 apple pies. If she wants to arrange all
    of the pies in rows of 5 pies each, how many rows will she have?
    \\\hline\hline

    \textbf{Solution}\qquad $(16 + 14) / 5 = 6$ \\\hline\hline
    \textbf{Math Concept needed for Each Operation} \\\hline
    \includegraphics[width=0.45\textwidth]{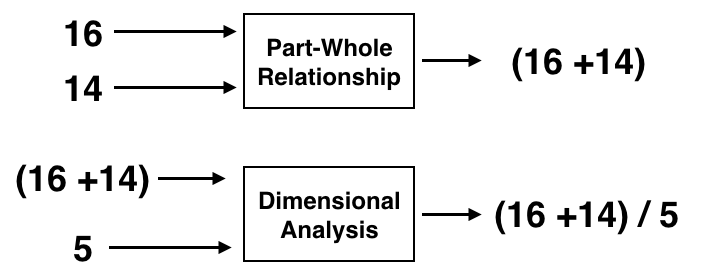} \\\hline

  \end{tabular}

  \caption{\small An example arithmetic word problem and its solution,
    along with the concepts required to generate each operation of the
    solution}

  \label{fig:running}
\end{figure}

Understanding and solving math word problems involves interpreting
natural language description of mathematical concepts, as well as
understanding their interaction with the physical world. Consider the
elementary school level arithmetic word problem shown in
Fig \ref{fig:running}.  To solve the problem, one needs to understand
that ``apple pies'' and ``pecan pies'' are kinds of ``pies'', and
hence, the number of apple pies and pecan pies needs to be summed up
to get the total number of pies.  Similarly, detecting that ``5''
represents ``the number of pies per row'' and applying dimensional
analysis or unit compatibility knowledge, helps us infer that the
total number of pies needs to be divided by $5$ to get the
answer. Besides part-whole relationship and dimensional analysis,
there are several other concepts that are needed to support reasoning
in math word problems. Some of these involve understanding
comparisons, transactions, and the application of math or physics
formulas. Most of this knowledge can be encoded as declarative rules,
as illustrated in this paper.

This paper introduces a framework for incorporating this ``declarative
knowledge'' into word problem solving. We focus on arithmetic word
problems, whose solution can be obtained by combining the numbers in
the problem with basic operations (addition, subtraction,
multiplication or division). For combining a pair of numbers or math
sub-expressions, our method first predicts the {\em math concept} that
is needed for it (e.g., subset relationship, dimensional analysis,
etc.), and then predicts a {\em declarative rule} under that concept
to infer the mathematical operation. We model the selection of
declarative rules as a latent variable, which removes the need for
expensive annotations for the intermediate steps.

The proposed approach has some clear advantages compared to existing
work on word problem solving. 
First, it provides interpretability of the solution, without expensive
annotations. Our method selects a declarative knowledge based
inference rule for each operation needed in the solution. These rules
provide an explanation for the operations performed. In particular, it
learns to select relevant rules without explicit annotations for them.
Second, each individual operation in the solution expression can
be generated independently by a separate mathematical concept. This
allows our method to handle multiple concepts in the same problem.

We show that existing datasets of arithmetic word problems suffer from
significant vocabulary biases and, consequently, existing solvers do
not do well on conceptually similar problems that are not biased in
the same way. Our method, on the other hand, learns the right
abstractions even in the presence of biases in the data. We also
introduce a novel approach to gather word problems without these
biases, creating a new dataset of $1492$ problems.

The next section discusses related work. We next introduce the
mathematical concepts required for arithmetic word problems, as well
as the declarative rules for each concept. Section \ref{sec:method}
describes our model -- how we predict answers using declarative
knowledge -- and provides the details of our training
paradigm. Finally, we provide experimental evaluation of our proposed
method in Section \ref{sec:exp}, and then conclude with a discussion
of future work.

\section{Related Work}
\label{sec:related}
Our work is primarily related to three major strands of research -
automatic word problem solving, semantic parsing, as well as 
approaches incorporating background knowledge in learning.

\subsection{Automatic Word Problem Solving}
There has been a growing interest in automatically solving math word
problems, with various systems focusing on particular types of
problems. These can be broadly categorized into two types: arithmetic
and algebra.

\noindent
\textbf{Arithmetic Word Problems} Arithmetic problems involve
combining numbers with basic operations (addition, subtraction,
multiplication and division), and are generally directed towards
elementary school students.  \newcite{RoyRo15}, \newcite{RoyRo17} as
well as this work focus on this class of word problems. The works of
\newcite{HHEK14} and \newcite{MitraBa16} focus on arithmetic problems
involving only addition and subtraction. Some of these approaches also
try to incorporate some form of declarative or domain
knowledge. \newcite{HHEK14} incorporates the transfer phenomenon by
classifying verbs; \newcite{MitraBa16} maps problems to a set of
formulas. Both require extensive annotations for intermediate steps
(verb classification for \newcite{HHEK14}, alignment of numbers to
formulas for \newcite{MitraBa16}, etc). In contrast, our method can
handle a more general class of problems, while training only requires
problem-equation pairs coupled with rate component annotations.
\newcite{RoyRo17} focuses only on using dimensional analysis knowledge,
and handles the same class of problems as we do. In contrast, our
method provides a framework for including any form of declarative
knowledge, exemplified here by incorporating common concepts required
for arithmetic problems.

\noindent
\textbf{Algebra Word Problems} Algebra word problems are characterized
by the use of (one or more) variables in contructing (one or more)
equations. These are typically middle or high school
problems. \newcite{KHSEA15} looks at single equation problems, and
\newcite{SWLLR15} focuses on number word problems. \newcite{KZBA14}
introduces a template based approach to handle general algebra word
problems and several works have later proposed improvements over this
approach \cite{ZhouDaCh15,UpChChYi16,HSLY17}. There has also been work
on generating rationale for word problem solving \cite{LYDB17}. More
recently, some focus turned to pre-university exam questions
\cite{MIIAA17,HPLLHJ17}, which requires handling a wider range of
problems and often more complex semantics.

\subsection{Semantic Parsing}
Our work is also related to learning semantic parsers from indirect
supervision \cite{CGCR10,LiangJoKl11}. The general approach here is to
learn a mapping of sentences to logical forms, with the only
supervision being the response of executing the logical form on a
knowledge base. Similarly, we learn to select declarative rules from
supervision that only includes the final operation (and not which rule
generated it). However, in contrast to the semantic parsing work, in
our case the selection of each declarative rule usually requires
reasoning across multiple sentences.  Also, we do not require an
explicit grounding of words or phrases to logical variables.

\subsection{Background Knowledge in Learning}
Approaches to incorporate knowledge in learning started with
Explanation based Learning (EBL) \cite{DeJong93,DeJong14}. EBL uses
domain knowledge based on observable predicates, whereas we learn to
map text to predicates of our declarative knowledge. More recent
approaches tried to incorporate knowledge in the form of constraints
or expectations from the output
\cite{RothYi04,ChangRaRo07,ChangRaRo12,GGGT10,SmithEi06,NCBJ10,BiskHo12,GimpelBa14}.

Finally, we note that there has been some work in the context of
Question Answering on perturbing questions or answers as a way to test
or assure the robustness of the approach or lack
of~\cite{KKSCER16,JiaLi17}. We make used of similar ideas in order to
generate an unbiased test set for Math word problems
(Sec.~\ref{sec:exp}).

\section{Knowledge Representation}
\label{sec:kr}
We introduce here our representation of domain knowledge. We organize
the knowledge hierarchically in two levels -- concepts and declarative
rules. A {\em math concept} is a phenomenon which needs to be
understood to apply reasoning over quantities. Examples of concepts
include part-whole relations, dimensional analysis, etc.  Under each
concept, there are a few declarative rules, which dictate which
operation is needed in a particular context. An example of a
declarative rule under {\em part-whole concept} can be that ``if two
numbers quantify ``parts" of a larger quantity, the operation between
them must be addition". These rules use concept specific predicates,
which we exemplify in the following subsections.

Since this work focuses on arithmetic word problems, we consider $4$
math concepts which are most common in these problems, as follows:
\begin{enumerate}
\item \textbf{Transfer}: This involves understanding the transfer of 
  objects from one person to another. For example, the action
  described by the sentence ``Tim gave 5 apples to Jim'', results in
  Tim losing ``5 apples'' and Jim gaining ``5 apples''.
\item \textbf{Dimensional Analysis}: This involves understanding
  compatibility of units or dimensions. For example, ``30
  pies'' can be divided by ``5 pies per row'' to get the number of
  rows.
\item \textbf{Part-Whole Relation}: This includes asserting that if
  two numbers quantify parts of a larger quantity, they are to be
  added. For example, the problem in Section \ref{sec:intro} involves
  understanding ``pecan pies'' and ``apple pies'' are parts of
  ``pies'', and hence must be added.
\item \textbf{Explicit Math}:  Word problems often mention explicit 
  math relationships among quantities or entities in the problem.  For
  example, ``Jim is 5 inches taller than Tim''. This concept captures
  the reasoning needed for such relationships.
\end{enumerate}

Each of these concepts comprises a small number of declarative rules
which determine the math operations; we describe them below.

\subsection{Transfer} 

Consider the following excerpt of a word problem exhibiting a transfer
phenomenon: {\em``Stephen owns $5$ books. Daniel gave him $4$ books.}
The goal of the declarative rules is to determine which operation is
required between $5$ and $4$, given that we know that a transfer is
taking place. We note that a transfer usually involves two entities,
which occur as subject and indirect object in a sentence. Also, the
direction of transfer is determined by the verbs associated with the
entities. We define a set of variables to denote these properties; we
define as Subj1, Verb1, IObj1 the subject, verb and indirect object
associated with the first number, and as Subj2, Verb2, IObj2 the
subject, verb and indirect object related to the second number. For
the above example, the assignment of the variables are shown below:
\begin{framed}{\noindent
[Stephen]$_{Subj1}$ [owns]$_{Verb1}$ $5$ books. [Daniel]$_{Subj2}$ 
[gave]$_{Verb2}$ [him]$_{IObj2}$ $4$ books.} 
\end{framed} 
In order to determine the direction of transfer, we require some
classification of verbs. In particular, we classify each verb into one
of five classes: HAVE, GET, GIVE, CONSTRUCT and DESTROY. The HAVE
class consists of all verbs which signify the state of an entity, such
as ``have'', ``own'', etc. The GET class contains verbs which indicate
the gaining of things for the subject. Examples of such verbs are
``acquire'', ``borrow'', etc. The GIVE class contains verbs which
indicate the loss of things for the subject.  Verbs like ``lend'',
``give'' belong to this class. Finally CONSTRUCT class constitutes
verbs indicating construction or creation, like ``build'', ``fill'',
etc., while DESTROY verbs indicate destruction related verbs like
``destroy'', ``eat'', ``use'', etc. This verb classification is
largely based on the work of \cite{HHEK14}.

Finally, the declarative rules for this concept have the
following form:
\begin{framed}{\noindent
$[\text{Verb1} \in \text{HAVE}] \wedge [\text{Verb2} \in \text{GIVE}] 
\wedge [\Coref(\text{Subj1}, \text{IObj2})] \Rightarrow \text{Addition}$}\end{framed}
\noindent
where $\Coref(A, B)$ is true when $A$ and $B$ represent the same
entity or are coreferent, and is false otherwise. In the examples
above, Verb1 is ``own'' and hence $[\text{Verb1} \in \text{HAVE}]$ is
true. Verb2 is ``give'' and hence $[\text{Verb2} \in \text{GIVE}]$ is
true. Finally, Subj1 and IObj2 both refer to Stephen, so
$[\Coref(\text{Subj1}, \text{IObj2})]$ returns true. As a result, the
above declarative rule dictates that addition should be performed
between $5$ and $4$.

We have $18$ such inference rules for transfer, covering all
combinations of verb classes and $\Coref()$ values. All these rules
generate addition or subtraction operations.

\subsection{Dimensional Analysis}
We now look at the use of dimensional analysis knowledge in word
problem solving. To use dimensional analysis, one needs to extract the
units of numbers as well as the relations between the units. Consider
the following excerpt of a word problem: {\em``Stephen has $5$
bags. Each bag has $4$ apples.} Knowing that the unit of $5$ is
``bag'' and the effective unit of $4$ is ``apples per bag'', allows us
to infer that the numbers can be multiplied to obtain the total number
of apples.

To capture these dependencies, we first introduce a few
terms. Whenever a number has a unit of the form ``A per B'', we refer
to ``A'' as the unit of the number, and refer to ``B'' as {\em the
rate component of the number}. In our example, the unit of $4$ is
``apple'', and the rate component of $4$ is ``bag''. We define
variables Unit1 and Rate1 to denote the unit and the rate component of
the first number respectively. We similarly define Unit2 and
Rate2. For the above example, the assignment of variables are shown
below:
\begin{framed}{\noindent
Stephen has $5$ [bags]$_{Unit1}$. Each [bag]$_{Rate2}$ has $4$
[apples]$_{Unit2}$.}\end{framed}
\noindent
Finally, the declarative rule applicable for our example has the
following form:
\begin{framed}{\noindent
$[\Coref(\text{Unit1}, \text{Rate2})] \Rightarrow \text{Multiplication}$}\end{framed}
\noindent
We only have $3$ rules for dimensional analysis. They generate
multiplication or division operations.

\subsection{Explicit Math}
In this subsection, we want to capture the reasoning behind explicit
math relationships expressed in word problems such as the one
described in: {\em``Stephen has $5$ apples. Daniel has $4$ more apples
than Stephen''}. 
We define by Math1 and Math2 any explicit math term associated with
the first and second numbers respectively. As was the case for
transfers, we also define Subj1, IObj1, Subj2, and IObj2 to denote the
entities participating in the math relationship. The assignment of
these variables in our example is:
\begin{framed}{\noindent
[Stephen]$_{Subj1}$ has $5$ apples. [Daniel]$_{Subj2}$ has $4$ [more
apples than]$_{Math2}$ [Stephen]$_{IObj2}$.}\end{framed}

We classify explicit math terms into one of three classes - ADD, SUB
and MUL. ADD comprises terms for addition, like ``more than'',
``taller than'' and ``heavier than''. SUB consists of terms for
subtraction like``less than'', ``shorter than'', etc., and MUL
contains terms indicating multiplication, like ``times'', ``twice''
and ``thrice''. Finally, the declarative rule that applies for our
example is:
\begin{framed}{\noindent
$[\Coref(\text{Subj1}, \text{IObj2})] \wedge [\text{Math2} \in \text{ADD}]
\Rightarrow \text{Addition}$}\end{framed}
\noindent
We have only $7$ rules for explicit math.

\subsection{Part-Whole Relation}
Understanding part-whole relationship entails understanding whether
two quantities are hyponym, hypernym or siblings (that is, co-hyponym,
or parts of the same quantity). For example, in the excerpt {\em
``Mrs. Hilt has 5 pecan pies and 4 apple pies''}, determining that
pecan pies and apple pies are parts of all pies, helps inferring that
addition is needed. We have 3 simple rules which directly map from
Hyponym, Hypernym or Sibling detection to the corresponding math
operation. For the above example, the applicable declarative rule is:
\begin{framed}{\noindent
$[\text{Sibling}(\text{Number1}, \text{Number2})] \Rightarrow \text{Addition}$}
\end{framed}
\noindent
The rules for part-whole concept can generate addition and subtraction
operations. 
Table \ref{tab:rules} gives a list of all the declarative rules. 
Note that all the declarative rules are designed to determine an
operation between two numbers only. We introduce a strategy in
Section \ref{sec:method}, which facilitates combining sub-expressions
with these rules.

\begin{table*}[!ht]
\centering
\begin{tabular}{|p{16cm}|}
\hline
Transfer \\\hline
$[\text{Verb1} \in \text{HAVE}] \wedge [\text{Verb2} \in \text{HAVE}] 
\wedge [\Coref(\text{Subj1}, \text{Subj2})] \Rightarrow -$\\

$[\text{Verb1} \in \text{HAVE}] \wedge [\text{Verb2} \in \text{(GET $\cup$ CONSTRUCT)}] 
\wedge [\Coref(\text{Subj1}, \text{Subj2})] \Rightarrow +$\\

$[\text{Verb1} \in \text{HAVE}] \wedge [\text{Verb2} \in \text{(GIVE $\cup$ DESTROY)}] 
\wedge [\Coref(\text{Subj1}, \text{Subj2})] \Rightarrow -$\\

$[\text{Verb1} \in \text{(GET $\cup$ CONSTRUCT)}] 
\wedge [\text{Verb2} \in \text{HAVE}]
\wedge [\Coref(\text{Subj1}, \text{Subj2})] \Rightarrow -$\\

$[\text{Verb1} \in \text{(GET $\cup$ CONSTRUCT)}] 
\wedge [\text{Verb2} \in \text{(GET $\cup$ CONSTRUCT)}]        
\wedge [\Coref(\text{Subj1}, \text{Subj2})] \Rightarrow +$\\

$[\text{Verb1} \in \text{(GET $\cup$ CONSTRUCT)}] 
\wedge [\text{Verb2} \in \text{(GIVE $\cup$ DESTROY)}]            
\wedge [\Coref(\text{Subj1}, \text{Subj2})] \Rightarrow -$\\

$[\text{Verb1} \in \text{(GIVE $\cup$ DESTROY)}] 
\wedge [\text{Verb2} \in \text{HAVE}]
\wedge [\Coref(\text{Subj1}, \text{Subj2})] \Rightarrow +$\\

$[\text{Verb1} \in \text{(GIVE $\cup$ DESTROY)}] 
\wedge [\text{Verb2} \in \text{(GET $\cup$ CONSTRUCT)}]        
\wedge [\Coref(\text{Subj1}, \text{Subj2})] \Rightarrow -$\\

$[\text{Verb1} \in \text{(GIVE $\cup$ DESTROY)}] 
\wedge [\text{Verb2} \in \text{(GIVE $\cup$ DESTROY)}]            
\wedge [\Coref(\text{Subj1}, \text{Subj2})] \Rightarrow +$\\\hline

We also have another rule for each rule above, which states that if
$\Coref(\text{Subj1}, \text{Obj2})$ or $\Coref(\text{Subj2},
\text{Obj1})$ is true, and none of the verbs is CONSTRUCT or DESTROY,
the final operation is changed from addition to subtraction, or vice
versa. \\\hline\hline

Dimensionality Analysis \\\hline
$[\Coref(\text{Unit1}, \text{Rate2}) \vee \Coref(\text{Unit2}, \text{Rate1})] 
\Rightarrow \times$ \\
$[\Coref(\text{Unit1}, \text{Unit2})] \wedge [\text{Rate2} \neq null]
 \Rightarrow \div$ \\
$[\Coref(\text{Unit1}, \text{Unit2})] \wedge [\text{Rate1} \neq null]
 \Rightarrow \div \text{ (Reverse order)}$ \\\hline\hline
 
 Explicit Math \\\hline
 $[\Coref(\text{Subj1}, \text{IObj2}) \vee \Coref(\text{Subj2}, \text{IObj1})] 
\wedge [\text{Math1} \in \text{ADD} \vee \text{Math2} \in \text{ADD}]    
\Rightarrow +$ \\

$[\Coref(\text{Subj1}, \text{IObj2}) \vee \Coref(\text{Subj2}, \text{IObj1})] 
\wedge [\text{Math1} \in \text{SUB} \vee \text{Math2} \in \text{SUB}]    
\Rightarrow -$

$[\Coref(\text{Subj1}, \text{Subj2})] 
\wedge [\text{Math1} \in \text{ADD} \vee \text{Math2} \in \text{ADD}]    
\Rightarrow -$

$[\Coref(\text{Subj1}, \text{Subj2})] 
\wedge [\text{Math1} \in \text{SUB} \vee \text{Math2} \in \text{SUB}]    
\Rightarrow +$

$[\Coref(\text{Subj1}, \text{Subj2})]                                                    \wedge [\text{Math1} \in \text{MUL}]                            
\Rightarrow \div\text{ (Reverse order)}$

$[\Coref(\text{Subj1}, \text{Subj2})]                                                    \wedge [\text{Math2} \in \text{MUL}]                            
\Rightarrow \div$

$[\Coref(\text{Subj1}, \text{IObj2}) \vee \Coref(\text{Subj2}, \text{IObj1})]        
\wedge [\text{Math1} \in \text{MUL} \vee \text{Math2} \in \text{MUL}]                        
\Rightarrow \times$ \\\hline\hline

Part-Whole Relationship \\\hline
$[\text{Sibling}(\text{Number1}, \text{Number2})] \Rightarrow +$\\
$[\text{Hyponym}(\text{Number1}, \text{Number2})] \Rightarrow -$\\
$[\text{Hypernym}(\text{Number1}, \text{Number2})] \Rightarrow -$\\\hline
\end{tabular}
\caption{\small List of declarative rules used in our system. $\div$ (reverse order)
indicates the second number being divided by the first. To determine the order of subtraction, we always subtract the smaller number from the larger number.}
\label{tab:rules}
\end{table*}

\section{Mapping of Word Problems to Declarative Knowledge}
\label{sec:method}
\begin{table*}[!ht] 
\centering
\begin{tabular}{|p{2.5cm}|p{13cm}|}\hline
Word Problem & Tim 's cat had 6 kittens . He gave 3 to Jessica. Then
Sara gave him 9 kittens . How many kittens does he now have ? \\\hline
\begin{center}Knowledge based Answer Derivation\end{center} & 
\vspace{-0.1in}
\includegraphics[width=0.8\textwidth]{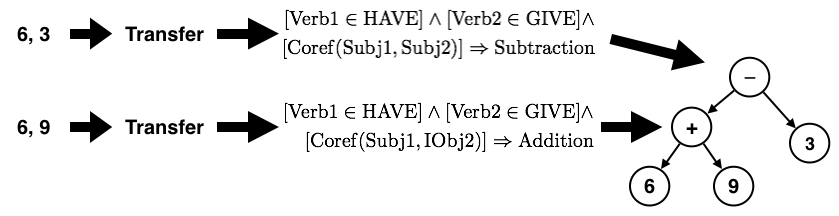}
\vspace{-0.25in}
\\\hline\hline 
Word Problem & Mrs. Hilt baked pies last weekend for a holiday
dinner. She baked 16 pecan pies and 14 apple pies. If she wants to
arrange all of the pies in rows of 5 pies each, how many rows will she
have? \\\hline
\begin{center}Knowledge based Answer Derivation\end{center} & 
\vspace{-0.1in}
\includegraphics[width=0.8\textwidth]{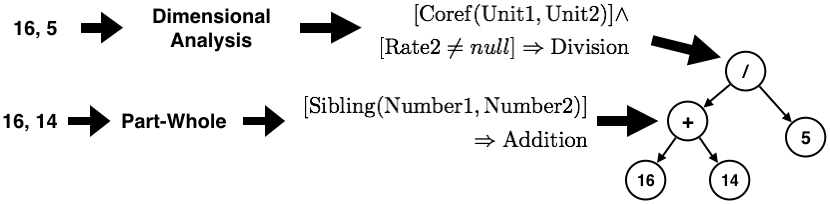}
\vspace{-0.25in}
\\\hline
\end{tabular}
\caption{\small Two examples of arithmetic word problems, and derivation of
  the answer. For each combination, first a math concept is chosen,
  and then a declarative rule from that concept is chosen to
  infer the operation. }
\label{fig:solving}
\end{table*}
Given an input arithmetic word problem $x$, the goal is to predict the
math expression $y$, which generates the correct answer. In order to
derive the expression $y$ from the word problem $x$, we leverage math
concepts and declarative rules that we introduced in Section
\ref{sec:kr}. In order to combine two numbers mentioned in $x$, we
first predict a concept $k$, and then we choose a declarative
knowledge rule $r$ from $k$. The rule $r$ generates the math operation
needed to combine the two numbers. Consider the first example in
Table~\ref{fig:solving}. To combine $6$ and $9$, we first decide on
the transfer concept, and then choose an appropriate rule under
transfer to generate the operation.

Next we need to combine the sub-expression $(6+9)$ with the number
$3$.  However, our inference rules were designed for the combination
of two numbers only. In order to combine a sub-expression, we choose
{\em a representative number} from the sub-expression, and use that
number to determine the operation. In our example, we choose the
number $6$ as the representative number for $(6+9)$, and decide the
operation between $6$ and $3$, following a similar procedure as
before. This operation is now used to combine $(6+9)$ and $3$.

The representative number for a sub-expression is chosen such
that it preserves the reasoning needed for the combination of this
sub-expression with other numbers. We follow a heuristic to
choose a representative number from a sub-expression:
\begin{enumerate}
\item For transfers and part-whole relationship, we choose the
  representative number of the left subtree.
\item In case of rate relationship, we choose the number which does
  not have a rate component.
\item In case of explicit math, we choose the number which is not
  directly associated with the explicit math expression.
\end{enumerate}

\subsection{Scoring Answer Derivations}

Given the input word problem $x$, the solution math expression $y$ is
constructed by combining numbers in $x$ with operations. We refer to
the set of operations used in an expression $y$ as $\Op(y)$. Each
operation $o$ in $\Op(y)$ is generated by first choosing a concept
$k^o$, and then selecting a declarative rule $r^o$ from that concept.

In order to discriminate between multiple candidate solution
expressions of a word problem $x$, we score them using a linear 
model over features extracted from the derivation of the solution.
Our scoring function has the following form:
\begin{align*}
\Score(x, y) = \sum_{o \in \Op(y)} w_{k}\phi_k(x, k^o) + w_{r}\phi_r(x, r^o)
\end{align*} 
where $\phi_k(x, k^o)$ and $\phi_r(x, r^o)$ are feature vectors
related to concept $k^o$, and declarative rule $r^o$,
respectively, and $w_k$ and $w_r$ are the corresponding weight
vectors. The term $w_{k}\phi_k(x, k^o)$ is the score for the selection
of $k^o$, and the term $w_{r}\phi_r(x, r^o)$ is the score for the
selection of $r^o$. Finally, the total score is the sum of the scores
of all concepts and rule choices, over all operations of $y$.

\subsection{Learning}

We wish to estimate the parameters of the weight vectors $w_k$ and
$w_r$, such that our scoring function assigns a higher score to the
correct math expression, and a lower score to other competing math
expressions.  For learning the parameters, we assume access to word
problems paired with the correct math expression. We show in Section
\ref{sec:details} that certain simple heuristics and rate component
annotations can be used to create somewhat noisy annotations for the
concepts needed for individual operations.  Hence, we will assume for
our formulation access to concept supervision as well. We thus assume
access to $m$ examples of the following form: $\{(x_1, y_1, \{k^o\}_{o
  \in \Op(y_1)}), (x_2, y_2, \{k^o\}_{o \in \Op(y_2)}),$ $\ldots,
(x_m, y_m, \{k^o\}_{o \in \Op(y_m)}) \}$.

We do not have any supervision for declarative rule selection, which
we model as a latent variable.

\noindent
\textbf{Two Stage Learning}: A straightforward solution for our
learning problem could be to jointly learn $w_k$ and $w_r$ using
latent structured SVM. However, we found that this model does not
perform well. Instead, we chose a two stage learning protocol. At the
first stage, we only learn $w_r$, the weight vector for scoring the
declarative rule choice. Once learned, we fix the parameters for
$w_r$, and then learn the parameters for $w_k$.

In order to learn the parameters for $w_r$, we solve:
{\small
\begin{align*}
\min_{w_r} \frac{1}{2}||w_r||^2 + C \sum_{i=1}^m \sum_{o \in \Op(y_i)} 
\Big[ \max_{\hat{r} \in k^o, \hat{r} \Rightarrow \hat{o}} 
w_r \cdot \phi_r(x, \hat{r}) + \\
\Delta(\hat{o}, o) \Big] - 
\max_{\hat{r} \in k^o, \hat{r} \Rightarrow o} 
w_r \cdot \phi_r(x, \hat{r}),\qquad\qquad
\end{align*}}
\noindent
where $\hat{r} \in k^o$ implies that $\hat{r}$ is a declarative rule
for concept $k^o$, $\hat{r} \Rightarrow o$ signify that the
declarative rule $\hat{r}$ generates operation $o$, and
$\Delta(\hat{o}, o)$ represents a measure of dissimilarity between
operations $o$ and $\hat{o}$. The above objective is similar to that
of latent structured SVM. For each operation $o$ in the solution
expression $y_i$, the objective tries to minimize the difference
between the highest scoring rule from its concept $k^o$, and highest
scoring rule from $k^o$ which explains or generates the operation $o$.

Next we fix the parameters of $w_r$, and solve:
{\small \begin{align*}
\min_{w_k} \frac{1}{2}||w_k||^2 + C \sum_{i=1}^m 
\qquad\qquad\qquad\qquad\\
\max_{y \in \mathcal{Y}} \left[ \Score(x_i, y) + \Delta(y, y_i) \right] 
- \Score(x_i, y_i).
\end{align*}}
\noindent
This is equivalent to a standard structured SVM objective. We use a
$0-1$ loss for $\Delta(\hat{o}, o)$. Note that fixing the parameters
of $w_r$ determines the scores for rule selection, removing the need
for any latent variables at this stage.

\subsection{Inference}
Given an input word problem $x$, inferring the best math expression
involves computing $\arg\max_{y \in \mathcal{Y}} \Score(x, y)$, where
$\mathcal{Y}$ is the set of all math expressions that can be created
by combining the numbers in $x$ with basic math operations.

The size of $\mathcal{Y}$ is exponential in the number of quantities
mentioned in $x$. As a result, we perform approximate inference using
beam search. We initialize the beam with the set $E$ of all numbers
mentioned in the problem $x$.  At each step of the beam search, we
choose two numbers (or sub-expressions) $e_1$ and $e_2$ from $E$, and
then select a math concept and a declarative rule to infer an
operation $o$. We create a new sub-expression $e_3$ by combining the
sub-expressions $e_1$ and $e_2$ with operation $o$. We finally create
a new set $E'$ from $E$, by removing $e_1$ and $e_2$ from it, and
adding $e_3$ to it. We remove $E$ from the beam, and add all such
modified sets $E'$ to the beam. We continue this process until all
sets in the beam have only one element in them. We choose the highest
scoring expression among these elements as the solution expression.

\section{Model and Implementation Details}
\label{sec:details}
\subsection{Supervision}

Each word problem in our dataset is annotated with the solution math
expression, along with alignment of numbers from the problem to the
solution expression. In addition, we also have annotations for the
numbers which possess a rate component. An example is shown in Fig 
\ref{fig:annotation}.
\begin{figure}[!ht]
\centering
\small
\begin{tabular}{|p{7.5cm}|}
\hline
\textbf{Problem:} Mrs. Hilt baked pies last weekend for a holiday
dinner. She baked 16 pecan pies and 14 apple pies. If she wants to
arrange all of the pies in rows of 5 pies each, how many rows will she
have? \\ \hline
\textbf{Number List:} $16$, $14$, $5$ \\ \hline
\textbf{Solution:} $(16_{[1]} + 14_{[2]}) / 5_{[3]} = 6$ \\ \hline
\textbf{Rates:} 5 \\ \hline
\end{tabular}
\caption{\small Annotations in our dataset. Number List refers to the
  numbers detected in the problem. The subscripts in the solution
  indicate the position of the numbers in the number list.}
\label{fig:annotation}
\end{figure}
This is the same level of supervision used in \cite{RoyRo17}. Many of
the annotations can be extracted semi-automatically. The number list
is extracted automatically by a number detector, the alignments
require human supervision only when the same numeric value is
mentioned multiple times in the problem. Most of the rate component
annotations can also be extracted automatically, see \cite{RoyRo17} for
details.

We apply a few heuristics to obtain noisy annotations for the
math concepts for operations. Consider the case for combining two
numbers $num1$ and $num2$, by operation $o$. We apply the following rules:
\begin{enumerate}
\item If we detect an explicit math pattern in the neighborhood of
  $num1$ or $num2$, we assign concept $k^o$ to be Explicit Math.
\item If $o$ is multiplication or division, and one of $num1$ or $num2$
  has a rate component, we assign $k^o$ to be Dimensional Analysis.
\item If $o$ is addition or subtraction, we check if the dependent
  verb of both numbers are identical. If they are, we assign $k^o$ to
  be Part-Whole relationship, otherwise, we assign it to be Transfer.
  We extract the dependent verb using the Stanford dependency parser
  \cite{ChenMa14}.
\end{enumerate}
The annotations obtained via these rules are of course not perfect. We
could not detect certain uncommon rate patterns like ``dividing the
cost $4$ ways'', and ``I read same number of books $4$ days running''.
There were part-whole relationships exhibited with complementary
verbs, as in ``I won $4$ games, and lost $3$.''. Both these cases lead
to noisy math concept annotations.

However, we tested a small sample of these annotations, and
found less than $5\%$ of them to be wrong. As a result, we assume
these annotations to be correct in our problem formulation.

\subsection{Features}
We use dependency parse labels and a small set of rules to extract
subject, indirect object, dependent verb, unit and rate component of
each number mentioned in the problem. Details of these extractions can
be found in the released codebase. Using these extractions, we define
two feature functions $\phi_k(x, k^o)$ and $\phi_r(x, r^o)$, where $x$
is the input word problem, and $k^o$ and $r^o$ are the concept and the
declarative rule for operation $o$ respectively. $\phi_r(x, r^o)$
constitutes the following features:
\begin{enumerate}
\item If $r^o$ contains $\Coref(\cdot)$ function, we add features
  related to similarity of the arguments of $\Coref(\cdot)$ (jaccard
  similarity score and presence of pronoun in one of the arguments).
\item For part-whole relationships, we add indicators for a list of
  words like ``remaining'', ``rest'', ``either'', ``overall'',
  ``total'', conjoined with the part-whole function in $r^o$ 
  (Hyponymy, Hypernymy, Sibling).
\item Unigrams from the neighborhood of numbers being combined.
\end{enumerate}
Finally, $\phi_k(x, k^o)$ generates the following features:
\begin{enumerate}
\item If $k^o$ is related to dimensional analysis, we add features
  indicating the presence of a rate component in the combining
  numbers.
\item If $k^o$ is part-whole, we add features indicating whether the
  verbs of combining numbers are identical.
\end{enumerate}
Note that these features capture several interpretable functions like
coreference, hyponymy, etc.

We do not learn three components of our system -- verb classification
for transfer knowledge, categorization of explicit math terms, and
irrelevant number detection. For verb classification, we use a seed
list of around $10$ verbs for each category. Given a new verb $v$, we
choose the most similar verb $v'$ from the seed lists according to
Glove vector \cite{glove:2014} based similarity .  We assign $v$ the
category of $v'$. This can be replaced by a learned component
\cite{HHEK14}. However we found seed list based categorization to work
well in most cases. For explicit math, we check for a small list of
patterns to detect and categorize math terms. Note that for both the
cases above, we still have to learn $\Coref(\cdot)$ function to
determine the final operation. Finally, to detect irrelevant numbers
(numbers which are not used in the solution), we use a set of rules
based on the units of numbers. Again, this can be replaced by a
learned model \cite{RoyRo15}.

\section{Experiments}
\label{sec:exp}
\subsection{Results on Existing Dataset}\label{subsec:early}
We first evaluate our approach on the existing datasets of AllArith,
AllArithLex, and AllArithTmpl \cite{RoyRo17}. AllArithLex and
AllArithTmpl are subsets of the AllArith dataset, created to test the
robustness to new vocabulary, and new equation forms respectively.  We
compare to the top performing systems for arithmetic word
problems. They are as follows:
\begin{enumerate}
\item $\Template$ : Template based algebra word problem solver of
  \cite{KZBA14}.
\item LCA++ : System of \cite{RoyRo15} based on lowest common
  ancestors of math expression trees.
\item $\UnitDep$: Unit dependency graph based solver of
  \cite{RoyRo17}.
\end{enumerate}  
We refer to our approach as $\Knowledge$. For all solvers, we use the
system released by the respective authors. The system of $\Template$
expects an equation as the answer, whereas our dataset contains only
math expressions. We converted expressions to equations by introducing
a single variable, and assigning the math expression to it.  For
example, an expression ``(2 + 3)'' gets converted to ``X = (2 + 3)''.

The first few columns of Table \ref{tab:full} shows the performance of
the systems on the aforementioned datasets\footnote{Results on the 
AllArith datasets are slightly different from \cite{RoyRo17}, since we 
fixed several ungrammatical sentences in the dataset}. The performance of
$\Knowledge$ is on par or lower than some of the existing systems.
We analyzed the systems, and found most of them to be not robust to
perturbations of the problem text; Table \ref{tab:bias} shows a few
examples. We further analyzed the datasets, and identified several
biases in the problems (in both train and test). Systems which
remember these biases get an undue advantage in evaluation. For
example, the verb ``give'' only appears with subtraction, and hence
the models are learning an erroneous correlation of ``give'' with
subtraction. Since the test also exhibit the same bias, these systems
get all the ``give''-related questions correct.  However, they fail to
solve the problem in Table \ref{tab:bias}, where ``give'' results in
addition.
\begin{table*}
  \centering
  \small
  \begin{tabular}{|l||p{1.5cm}|p{1.5cm}|p{1.5cm}||p{1.5cm}|p{1.5cm}|p{1.5cm}||p{1.5cm}||}
    \hline
    System & AllArith & AllArith Lex & AllArith Tmpl &
    Aggregate & Aggregate Lex & Aggregate Tmpl &
    Train on AllArith, Test on Perturb\\\hline\hline
    $\Template$ & 71.96 & 64.09 & 70.64 & 54.62 & 45.05 & 54.69 & 24.2 \\\hline
    LCA++ & 78.34 & 66.99 & 75.66 & 65.21 & 53.62 & 63.0 & 43.57 \\\hline
    $\UnitDep$ & \textbf{79.67} & 71.33 & \textbf{77.11} & 69.9 & 57.51 & 
    \textbf{68.64} & 46.29\\\hline
    $\Knowledge$ & 77.86 & \textbf{72.53} & 74.7 & \textbf{73.32$^*$} & 
    \textbf{66.63$^*$} & 68.62 & \textbf{65.66$^*$}\\\hline
  \end{tabular}
  \caption{\small Accuracy in solving arithmetic word problems.
  All columns except the last report 5-fold cross validation
    results. $^*$ indicates statistically significant improvement
    ($p=0.05$) over second highest score in the column.  }
  \label{tab:full}
\end{table*}
\begin{table*}
\centering
\small
\begin{tabular}{|p{7.1cm}|p{3.9cm}|p{3.9cm}|}\hline
\multirow{2}{*}{Problem} & \multicolumn{2}{p{7.8cm}|}{Systems which solved correctly} \\\cline{2-3}
& Trained on AllArith & Trained on Aggregate \\\hline\hline
Adam has 70 marbles. Adam gave 27 marbles to Sam. How many marbles does Adam have now?
&  $\Template$, $\UnitDep$, LCA, $\Knowledge$ & LCA, $\UnitDep$, $\Knowledge$ \\\hline
Adam has 70 marbles. Sam gave 27 marbles to Adam. How many marbles does Adam have now?
& $\Knowledge$ & $\Template$, $\Knowledge$ \\\hline\hline
Adam has 5 marbles. Sam has 6 more marbles than Adam. How many marbles does Sam have?
& LCA, $\UnitDep$, $\Knowledge$ & LCA, $\UnitDep$, $\Knowledge$ \\\hline
Adam has 11 marbles. Adam has 6 more marbles than Sam. How many marbles does Sam have?
& $\Template$, $\Knowledge$ & $\Template$, $\Knowledge$ \\\hline
\end{tabular}
\caption{\small Pairs of pertubed problems, along with the systems
  which get them correct}
\label{tab:bias}
\end{table*}
We also tested $\Knowledge$ on the addition subtraction problems
dataset released by \cite{HHEK14}. It achieved a cross validation
accuracy of $77.19\%$, which is competitive with the state of the art
accuracy of $78\%$ achieved with the same level of supervision. The
system of \cite{MitraBa16} achieved $86.07\%$ accuracy on this
dataset, but requires rich annotations for formulas and alignment of
numbers to formulas.

\subsection{New Dataset Creation}

In order to remove the aforementioned biases from the dataset, we
augment it with new word problems collected via a crowdsourcing
platform. These new word problems are created by perturbing the
original problems minimally, such that the answer is different from
the original problem. For each word problem $p$ with an answer
expression $a$ in our original dataset AllArith, we replace one
operation in $a$ to create a new math expression $a'$. We ask
annotators to modify problem $p$ minimally, such that $a'$ is now the
solution to the modified word problem.

We create $a'$ from $a$ either by replacing an addition with
subtraction or vice versa, or by replacing multiplication with
division or vice versa. We do not replace addition and subtraction
with multiplication or division, since there might not be an easy
perturbation that supports this conversion. We only allowed perturbed
expressions which evaluate to values greater than $1$. For example, we
generate the expression ``(3+2)'' from ``(3-2)'', we generated
expressions ``(10+2)/4'' and ``(10-2)*4'' for the expression
``(10-2)/4''. We generate all possible perturbed expressions for a
given answer expression, and ask for problem text modification for
each one of them.

We show the annotators the original problem text $p$ paired with a
perturbed answer $a'$. The instructions advised them to copy over the
given problem text, and modify it as little as possible so that the
given math expression is now the solution to this modified
problem. They were also instructed to not add or delete the numbers
mentioned in the problem. If the original problem mentions two
``3''s and one ``2'', the modified problem should also contain two
``3''s and one ``2''.

We manually pruned problems which did not yield the desired solution
$a'$, or were too different from the input problem $p$.  This
procedure gave us a set of $661$ new word problems, which we refer to
as \textbf{Perturb}. Finally we augment \textbf{AllArith} with the
problems of \textbf{Perturb}, and call this new dataset
\textbf{Aggregate}. Aggregate has a total of $1492$ problems.

The addition of the \textbf{Perturb} problems ensures that the dataset
now has problems with similar lexical items generating different
answers. This minimizes the bias that we discussed in subsection
\ref{subsec:early}. To quantify this, consider the probability
distribution over operations for a quantity $q$, given that word $w$
is present in the neighborhood of $q$. For an unbiased dataset, you
will expect the entropy of this distribution to be high, since the
presence of a single word in a number neighborhood will seldom be
completely informative for the operation. We compute the average of
this entropy value over all numbers and neighborhood words in our
dataset.  AllArith and Perturb have an average entropy of $0.34$ and
$0.32$ respectively, whereas Aggregate's average entropy is $0.54$,
indicating that, indeed, the complete data set is significantly less
biased.

\subsection{Generalization from Biased Dataset}
First, we evaluate the ability of systems to generalize from biased
datasets. We train all systems on AllArith, and test them on Perturb
(which was created by perturbing AllArith problems). The last column
of Table \ref{tab:full} shows the performance of systems in this
setting. $\Knowledge$ outperforms all other systems in this setting
with around $19 \%$ absolute improvement over $\UnitDep$. This shows
that declarative knowledge allows the system to learn the correct
abstractions, even from biased datasets.

\subsection{Results on the New Dataset}
Finally, we evaluate the systems on the Aggregate dataset. Following
previous work \cite{RoyRo17}, we compute two subsets of Aggregate
comprising $756$ problems each, using the MAWPS \cite{KRAKH16}
system. The first, called AggregateLex, is one with low lexical
repetitions, and the second called AggregateTmpl is one with low
repetitions of equation forms. We also evaluate on these two subsets
on a 5-fold cross-valiation. Columns 4-6 of Table \ref{tab:full} show
the performance of systems on this setting.  $\Knowledge$
significantly outperforms other systems on Aggregate and AggregateLex,
and is similar to $\UnitDep$ on AggregateTmpl. There is a $9 \%$
absolute improvement on AggregateLex, showing that $\Knowledge$ is
significantly more robust to low lexical overlap between train and
test. The last column of Table \ref{tab:bias} also shows that the
other systems do not learn the right abstraction, even when trained on
Aggregate.

\subsection{Analysis}
\noindent
\textbf{Coverage of the Declarative Rules} We chose math concepts and
declarative rules based on their prevalance in arithmetic word
problems. We found that the four concepts introduced in this paper
cover almost all the problems in our dataset; only missing $4$
problems involving application of area formulas. We also checked
earlier arithmetic problem datasets from the works of
\cite{HHEK14,RoyRo15}, and found that the math concepts and
declarative rules introduced in this paper cover all their problems.

A major challenge in applying these concepts and rules to algebra word
problems is the use of variables in constructing equations. Variables
are often implicitly described, and it is difficult to extract units,
dependent verbs, associated subjects and objects for the variables.
However, we need these extractions in order to apply our declarative
rules to combine variables. There has been some work to extract
meaning of variables \cite{RoyUpRo16} in algebra word problems; an
extension of this can possibly support the application of rules in
algebra word problems. We leave this exploration to future work. 

Higher standard word problems often require application of math
formulas like ones related to area, interest, probability,
etc. Extending our approach to handle such problems will involve
encoding math formulas in terms of concepts and rules, as well as
adding concept specific features to the learned predictors. The
declarative rules under the Explicit Math category currently handles
simple cases, this set needs to be augmented to handle complex
number word problems found in algebra datasets.

\noindent
\textbf{Gains achieved by Declarative Rules}
Table \ref{tab:wins} shows examples of problems which $\Knowledge$
gets right, but $\UnitDep$ does not. The gains can be attributed to
the injection of declarative knowledge. Earlier systems
like $\UnitDep$ try to learn the reasoning required for these problems
from the data alone. This is often difficult in the presence of
limited data, and noisy output from NLP tools. In contrast, we learn
probabilistic models for interpretable functions like coreference,
hyponymy, etc., and then use declarative knowledge involving these
functions to perform reasoning. This considerably reduces the
complexity of the target function to be learnt, and hence we end up
with a more robust model.

\noindent
\textbf{Effect of Beam Size}
We used a beam size of $1000$ in all our experiments. However, we found
that varying the beam size does not effect the performance significantly.
Even lowering the beam size to $100$ reduced performance by only $1\%$.

\noindent
\textbf{Weakness of Approach}
A weakness of our method is the requirement to have all relevant
declarative knowledge during training.  Many of the component
functions (like coreference) are learnt through latent alignments with
no explicit annotations. If too many problems are not explained by the
knowledge, the model will learn noisy alignments for the component
functions.
 
\begin{table}
\small
\centering
\begin{tabular}{|p{6.9cm}|}\hline

Isabel had 2 pages of math homework and 4 pages of reading
homework. If each page had 5 problems on it, how many problems did she
have to complete total ? \\\hline

Tim's cat had kittens. He gave 3 to Jessica and 6 to Sara . He now has
9 kittens . How many kittens did he have to start with ? \\\hline

Mrs. Snyder made 86 heart cookies. She made 36 red cookies, and
the rest are pink. How many pink cookies did she make? \\\hline

\end{tabular}
\caption{\small Examples which $\Knowledge$ gets correct, but
  $\UnitDep$ does not.}
\label{tab:wins}
\end{table}

Table \ref{tab:errors} shows the major categories of errors with
examples. $26\%$ of the errors are due to extraneous number detection.
We use a set of rules based on units of numbers, to detect such
irrelevant numbers. As a result, we fail to detect numbers which are
irrelevant due to other factors, like associated entities, or
associated verb. We can potentially expand our rule based system to
detect those, or replace it by a learned module like
\cite{RoyRo15}. Another major source of errors is parsing of rate
components, that is, understanding ``earns \$46 cleaning a home''
should be normalized to ``46\$ per home''. Although we learn a model
for coreference function, we make several mistakes related to
coreference. For the example in Table \ref{tab:errors}, we fail to
detect the coreference between ``team member'' and ``people''.

\begin{table}
\small
\centering
\begin{tabular}{|p{1.9cm}|p{5cm}|}\hline

Irrelevant Number Detection (26\%)& Sally had 39 baseball cards, and
\underline{9 were torn}. Sara bought 24 of Sally's baseball cards
. How many baseball cards does Sally have now?\\\hline

Parsing Rate Component (26\%)& Mary earns \underline{\$46 cleaning a
  home}. How many homes did she clean, if she made 276
dollars?\\\hline

Coreference (22\%)& There are 5 \underline{people} on the Green Bay
High track team. If a relay race is 150 meters long, how far will each
\underline{team member} have to run? \\\hline

\end{tabular}
\caption{\small Examples of errors made by $\Knowledge$}
\label{tab:errors}
\end{table}

\section{Conclusion}
\label{sec:conclusion}
In this paper, we introduce a framework for incorporating declarative
knowledge in word problem solving. Our knowledge based approach
outperforms all other systems, and also learns better abstractions
from biased datasets. Given that the variability in text is much
larger than the number of declarative rules that governs Math word
problems, we believe that this is a good way to introduce Math
knowledge to a natural language understanding system. Consequently,
future
work will involve extending our approach to handle a wider
range of word problems, possibly by supporting better grounding of
implicit variables and including a larger number of math concepts and
declarative rules. An orthogonal exploration direction is to apply
these techniques to generate summaries of financial or sports news, or
generate statistics of war or gun violence deaths from news corpora. A
straightforward approach can be to augment news documents with a
question asking for the required information, and treating this
augmented news document as a math word problem.

Code and dataset are available at \url{https://github.com/CogComp/arithmetic}.

\section*{Acknowledgments}
This work is funded by DARPA under agreement number
FA8750-13-2-0008, and a grant from the Allen Institute for
Artificial Intelligence (allenai.org).

\bibliography{ccg,cited,subhro}

\begin{thebibliography}{}

\bibitem[\protect\citename{Bisk and Hockenmaier}2012]{BiskHo12}
Yonatan Bisk and Julia Hockenmaier.
\newblock 2012.
\newblock {Simple Robust Grammar Induction with Combinatory Categorial
  Grammars}.
\newblock In {\em Proceedings of the Twenty-Sixth Conference on Artificial
  Intelligence (AAAI-12)}, pages 1643--1649, Toronto, Canada, July.

\bibitem[\protect\citename{Chang \bgroup et al.\egroup }2012]{ChangRaRo12}
Ming-Wei Chang, Lev Ratinov, and Dan Roth.
\newblock 2012.
\newblock Structured learning with constrained conditional models.
\newblock {\em Machine Learning}, 88(3):399--431, 6.

\bibitem[\protect\citename{Chen and Manning}2014]{ChenMa14}
Danqi Chen and Christopher~D Manning.
\newblock 2014.
\newblock A fast and accurate dependency parser using neural networks.
\newblock In {\em Empirical Methods in Natural Language Processing (EMNLP)}.

\bibitem[\protect\citename{Clarke \bgroup et al.\egroup }2010]{CGCR10}
James Clarke, Dan Goldwasser, Ming-Wei Chang, and Dan Roth.
\newblock 2010.
\newblock Driving semantic parsing from the world's response.
\newblock In {\em Proc. of the Conference on Computational Natural Language
  Learning (CoNLL)}, 7.

\bibitem[\protect\citename{DeJong}1993]{DeJong93}
Gerald DeJong.
\newblock 1993.
\newblock {\em Investigating explanation-based learning}.
\newblock Kluwer international series in engineering and computer science.
  Kluwer Academic Publishers.

\bibitem[\protect\citename{DeJong}2014]{DeJong14}
Gerald DeJong.
\newblock 2014.
\newblock Explanation-based learning.
\newblock In T.~Gonzalez, J.~Diaz-Herrera, and A.~Tucker, editors, {\em CRC
  computing handbook: Computer science and software engineering}, pages
  66.1--66.26. CRC Press, Boca Raton.

\bibitem[\protect\citename{Ganchev \bgroup et al.\egroup }2010]{GGGT10}
Kuzman Ganchev, Joao Gra\c{c}a, Jennifer Gillenwater, and Ben Taskar.
\newblock 2010.
\newblock Posterior regularization for structured latent variable models.
\newblock {\em Journal of Machine Learning Research}.

\bibitem[\protect\citename{Gimpel and Bansal}2014]{GimpelBa14}
Kevin Gimpel and Mohit Bansal.
\newblock 2014.
\newblock Weakly-supervised learning with cost-augmented contrastive
  estimation.
\newblock In {\em Proc. of EMNLP}.

\bibitem[\protect\citename{Hopkins \bgroup et al.\egroup }2017]{HPLLHJ17}
Mark Hopkins, Cristian Petrescu-Prahova, Roie Levin, Ronan Le~Bras, Alvaro
  Herrasti, and Vidur Joshi.
\newblock 2017.
\newblock Beyond sentential semantic parsing: Tackling the math sat with a
  cascade of tree transducers.
\newblock In {\em Proceedings of the 2017 Conference on Empirical Methods in
  Natural Language Processing}, pages 806--815, Copenhagen, Denmark, September.
  Association for Computational Linguistics.

\bibitem[\protect\citename{Hosseini \bgroup et al.\egroup }2014]{HHEK14}
Mohammad~Javad Hosseini, Hannaneh Hajishirzi, Oren Etzioni, and Nate Kushman.
\newblock 2014.
\newblock Learning to solve arithmetic word problems with verb categorization.
\newblock In {\em Proceedings of the Conference on Empirical Methods for
  Natural Language Processing (EMNLP)}.

\bibitem[\protect\citename{Huang \bgroup et al.\egroup }2017]{HSLY17}
Danqing Huang, Shuming Shi, Chin-Yew Lin, and Jian Yin.
\newblock 2017.
\newblock Learning fine-grained expressions to solve math word problems.
\newblock In {\em Proceedings of the 2017 Conference on Empirical Methods in
  Natural Language Processing}, pages 816--825, Copenhagen, Denmark, September.
  Association for Computational Linguistics.

\bibitem[\protect\citename{Jia and Liang}2017]{JiaLi17}
Robin Jia and Percy Liang.
\newblock 2017.
\newblock Adversarial examples for evaluating reading comprehension systems.
\newblock In {\em Proceedings of the 2017 Conference on Empirical Methods in
  Natural Language Processing}, pages 2021--2031. Association for Computational
  Linguistics, September.

\bibitem[\protect\citename{Khashabi \bgroup et al.\egroup }2016]{KKSCER16}
Daniel Khashabi, Tushar Khot, Ashish Sabharwal, Peter Clark, Oren Etzioni, and
  Dan Roth.
\newblock 2016.
\newblock Question answering via integer programming over semi-structured
  knowledge.
\newblock In {\em Proc. of the International Joint Conference on Artificial
  Intelligence (IJCAI)}.

\bibitem[\protect\citename{Koncel-Kedziorski \bgroup et al.\egroup
  }2015]{KHSEA15}
Rik Koncel-Kedziorski, Hannaneh Hajishirzi, Ashish Sabharwal, Oren Etzioni, and
  Siena Ang.
\newblock 2015.
\newblock {P}arsing {A}lgebraic {W}ord {P}roblems into {E}quations.
\newblock {\em TACL}.

\bibitem[\protect\citename{Koncel-Kedziorski \bgroup et al.\egroup
  }2016]{KRAKH16}
Rik Koncel-Kedziorski, Subhro Roy, Aida Amini, Nate Kushman, and Hannaneh
  Hajishirzi.
\newblock 2016.
\newblock Mawps: A math word problem repository.
\newblock In {\em NAACL}.

\bibitem[\protect\citename{Kushman \bgroup et al.\egroup }2014]{KZBA14}
Nate Kushman, Luke Zettlemoyer, Regina Barzilay, and Yoav Artzi.
\newblock 2014.
\newblock Learning to automatically solve algebra word problems.
\newblock In {\em Proceedings of the Annual Meeting of the Association for
  Computational Linguistics (ACL)}, pages 271--281.

\bibitem[\protect\citename{Liang \bgroup et al.\egroup }2011]{LiangJoKl11}
Percy Liang, Michael Jordan, and Dan Klein.
\newblock 2011.
\newblock Learning dependency-based compositional semantics.
\newblock In {\em Proceedings of the Annual Meeting of the Association for
  Computational Linguistics (ACL)}.

\bibitem[\protect\citename{Ling \bgroup et al.\egroup }2017]{LYDB17}
Wang Ling, Dani Yogatama, Chris Dyer, and Phil Blunsom.
\newblock 2017.
\newblock Program induction by rationale generation: Learning to solve and
  explain algebraic word problems.
\newblock In {\em ACL}.

\bibitem[\protect\citename{Matsuzaki \bgroup et al.\egroup }2017]{MIIAA17}
Takuya Matsuzaki, Takumi Ito, Hidenao Iwane, Hirokazu Anai, and Noriko H.~Arai.
\newblock 2017.
\newblock Semantic parsing of pre-university math problems.
\newblock In {\em Proceedings of the 55th Annual Meeting of the Association for
  Computational Linguistics (Volume 1: Long Papers)}, pages 2131--2141,
  Vancouver, Canada, July. Association for Computational Linguistics.

\bibitem[\protect\citename{Mitra and Baral}2016]{MitraBa16}
Arindam Mitra and Chitta Baral.
\newblock 2016.
\newblock Learning to use formulas to solve simple arithmetic problems.
\newblock In {\em ACL}.

\bibitem[\protect\citename{Naseem \bgroup et al.\egroup }2010]{NCBJ10}
Tahira Naseem, Harr Chen, Regina Barzilay, and Mark Johnson.
\newblock 2010.
\newblock Using universal linguistic knowledge to guide grammar induction.
\newblock In {\em Proceedings of the 2010 Conference on Empirical Methods in
  Natural Language Processing}, EMNLP '10, pages 1234--1244, Stroudsburg, PA,
  USA. Association for Computational Linguistics.

\bibitem[\protect\citename{Pennington \bgroup et al.\egroup }2014]{glove:2014}
Jeffrey Pennington, Richard Socher, and Christopher~D. Manning.
\newblock 2014.
\newblock Glove: Global vectors for word representation.
\newblock In {\em Proc. of EMNLP}.

\bibitem[\protect\citename{Roth and tau Yih}2004]{RothYi04}
Dan Roth and Wen tau Yih.
\newblock 2004.
\newblock A linear programming formulation for global inference in natural
  language tasks.
\newblock In Hwee~Tou Ng and Ellen Riloff, editors, {\em Proc. of the
  Conference on Computational Natural Language Learning (CoNLL)}, pages 1--8.
  Association for Computational Linguistics.

\bibitem[\protect\citename{Roy and Roth}2015]{RoyRo15}
Subhro Roy and Dan Roth.
\newblock 2015.
\newblock Solving general arithmetic word problems.
\newblock In {\em Proc. of the Conference on Empirical Methods in Natural
  Language Processing (EMNLP)}.

\bibitem[\protect\citename{Roy and Roth}2017]{RoyRo17}
Subhro Roy and Dan Roth.
\newblock 2017.
\newblock Unit dependency graph and its application to arithmetic word problem
  solving.
\newblock In {\em Proc. of the Conference on Artificial Intelligence (AAAI)}.

\bibitem[\protect\citename{Roy \bgroup et al.\egroup }2016]{RoyUpRo16}
Subhro Roy, Shyam Upadhyay, and Dan Roth.
\newblock 2016.
\newblock Equation parsing : Mapping sentences to grounded equations.
\newblock In {\em Proc. of the Conference on Empirical Methods in Natural
  Language Processing (EMNLP)}.

\bibitem[\protect\citename{Shi \bgroup et al.\egroup }2015]{SWLLR15}
Shuming Shi, Yuehui Wang, Chin-Yew Lin, Xiaojiang Liu, and Yong Rui.
\newblock 2015.
\newblock Automatically solving number word problems by semantic parsing and
  reasoning.
\newblock In {\em EMNLP}.

\bibitem[\protect\citename{Smith and Eisner}2006]{SmithEi06}
Noah Smith and Jason Eisner.
\newblock 2006.
\newblock Annealing structural bias in multilingual weighted grammar induction.
\newblock In {\em Proceedings of the Annual Meeting of the Association for
  Computational Linguistics (ACL)}, ACL-44, pages 569--576, Stroudsburg, PA,
  USA. Association for Computational Linguistics.

\bibitem[\protect\citename{Upadhyay \bgroup et al.\egroup }2016]{UpChChYi16}
Shyam Upadhyay, Ming-Wei Chang, Kai-Wei Chang, and Wen-tau Yih.
\newblock 2016.
\newblock Learning from explicit and implicit supervision jointly for algebra
  word problems.
\newblock In {\em EMNLP}.

\bibitem[\protect\citename{wei Chang \bgroup et al.\egroup }2007]{ChangRaRo07}
Ming wei Chang, Lev Ratinov, and Dan Roth.
\newblock 2007.
\newblock Guiding semi-supervision with constraint-driven learning.
\newblock In {\em Proc. of the Annual Meeting of the Association for
  Computational Linguistics (ACL)}, pages 280--287, Prague, Czech Republic, 6.
  Association for Computational Linguistics.

\bibitem[\protect\citename{Zhou \bgroup et al.\egroup }2015]{ZhouDaCh15}
Lipu Zhou, Shuaixiang Dai, and Liwei Chen.
\newblock 2015.
\newblock Learn to solve algebra word problems using quadratic programming.
\newblock In {\em EMNLP}.

\end{thebibliography}
\bibliographystyle{acl2012}
\end{document}